\documentclass[sigconf,nonacm]{acmart}
\usepackage{graphicx} 
\usepackage{color}
\usepackage{xcolor}
\usepackage{colortbl}

\definecolor{green}{rgb}{0.1,0.7,0.2}
\newcommand{\grayline}{\arrayrulecolor{gray}\hline\arrayrulecolor{black}}

\title{Evolving Financial Trading Strategies\\ with Vectorial Genetic Programming}

\author{Rui Menoita}
\affiliation{%
	\institution{LASIGE, Faculdade de Ciências,\\Universidade de Lisboa}
    \country{Portugal}
}
\orcid{0009-0000-6046-9992}
\author{Sara Silva}
\affiliation{%
	\institution{LASIGE, Faculdade de Ciências,\\Universidade de Lisboa}
    \country{Portugal}
}
\email{sara@fc.ul.pt}
\orcid{0000-0001-8223-4799}

\begin{document}

\begin{abstract}
Establishing profitable trading strategies in financial markets is a challenging task. While traditional methods like technical analysis have long served as foundational tools for traders to recognize and act upon market patterns, the evolving landscape has called for more advanced techniques. We explore the use of Vectorial Genetic Programming (VGP) for this task, introducing two new variants of VGP, one that allows operations with complex numbers and another that implements a strongly-typed version of VGP. We evaluate the different variants on three financial instruments, with datasets spanning more than seven years. Despite the inherent difficulty of this task, it was possible to evolve profitable trading strategies. A comparative analysis of the three VGP variants and standard GP revealed that standard GP is always among the worst whereas strongly-typed VGP is always among the best.
\end{abstract}

\maketitle

\section{Introduction} \label{sect:introduction}

Financial time series forecasting poses significant challenges, as underscored by some economists through the efficient market hypothesis~\cite{10.2307/2325486}. It suggests that if market participants could forecast the price, they could generate unlimited profits, and this profit-seeking behavior would destroy any pattern of predictability that might emerge in the series. This opens the question of whether the future market price is even predictable. However, a considerable number of studies in the stock market~\cite{Bollerslev,CAMPBELL200627,FAMA1983,GOLEZ2018248,Liu2020-mz} and in the forex (foreign exchange) market~\cite{Neely1997} have shown that some predictability arises and can be exploited to obtain profits.

Technical analysis is a method that attempts to exploit these patterns. Professional traders use it to extract useful information from the asset price variance, using information about the historical price movements in charts and formulas to forecast future prices and trends. These investors argue that this method allows them to profit from changes in market psychology, a view that is well expressed by Pring~\cite{pring_2014}. Technical analysis was initially designed for stock markets but quickly expanded and is now used in various markets such as forex and cryptocurrency.

We propose a system that uses the knowledge of technical analysis indicators and the learning ability of evolutionary algorithms to generate profitable trading rules. In particular, Genetic Programming (GP) is used~\cite{koza1992,poli08} in one of its newest formulations called Vectorial GP (VGP), introduced by Azzali et al.~\cite{azzali2019vectorial}. Vectorial GP extends standard GP by allowing the use of vectors as input data and introducing new operators that work on these vectors. This new approach gives more context to the GP agent to work with, which means that the agent will be aware of past data to make a more informed decision, hopefully resulting in better performance as a forecasting model. Vectorial GP has shown promising results, outperforming standard GP and some standard machine learning techniques in different applications~\cite{azzali2019vectorial,Azzali2020,Azzali2020TowardsTU,fleck2022grammar,Abbona2022TowardsAV,fleck2023vectorial,Azzali2024AutomaticFE}. 
Different variants of the original VGP have also been proposed~\cite{Azzali2020InvestigatingTU,Fleck2021GrammarBasedVG,Gligorovski2023LGPVECAV}.

Starting from the original VGP formulation, we also propose new variants of the original method and study their ability to evolve profitable trading strategies in different financial market scenarios. In particular, we introduce a VGP variant that allows operations with complex numbers and another variant that implements a strongly-typed version of VGP. In the remainder of this document, we explain the basics of financial markets, describe previous related work, explain the data and methods used, specify our experiments, discuss the results and draw conclusions.

\section{Background} \label{sect:background}

The cornerstone of financial market analysis lies in forecasting price behaviors, which in turn facilitates informed decisions during the purchase or sale of financial assets. Market analysis predominantly bifurcates into two categories: technical analysis and fundamental analysis. Although these two forms of analysis can be used in conjunction to achieve optimal conclusions in the financial market, it is important to acknowledge their distinct areas of focus.

Technical analysis~\cite{Murphy} is a trading discipline that involves examining statistical trends derived from trading activity, such as price movement and volume, to assess investments and identify potential trading opportunities. Technical analysts focus on price movement patterns, trading signals, and other analytical charting tools to evaluate the strength or weakness of an instrument, in contrast to fundamental analysts who attempt to determine an instrument's underlying value based on financial or economic data~\cite{Elder}.

The versatility of technical analysis allows its application across a gamut of instruments, provided they possess historical trading data. This spans stocks, futures, commodities, and currencies, among other assets. However, it is more commonly used in commodities and forex markets, where traders place a greater emphasis on short-term price changes.

\subsection{Technical Indicators}\label{subsect:indicators}

Technical indicators, grounded in heuristics or pattern-based signals, emerge from the analysis of factors like price, volume, and, in some cases, open interest associated with an asset or contract. They fall into two primary categories: classic and computational. Classic indicators rely on graphical patterns (normally using the candlestick chart, Figure~\ref{fig:candle_chart}), while computational indicators use formulas and numerical calculations. Computational technical indicators can be categorized into four groups~\cite{Elder,Murphy}:
\begin{itemize}
\item \textbf{Trend followers:} These indicators measure the direction and strength of trends. They are typically represented as lines on a chart and provide sell or buy signals when the price is below or above the indicator, respectively. The slope of the line can also indicate the strength of the trend. Examples of trend follower indicators include Simple Moving Average (SMA), Exponential Moving Average (EMA), and Hull Moving Average.
\item \textbf{Oscillators:} This type of indicator oscillates between two values and helps measure the strength and momentum of a trend. Oscillators often signal whether the market is overbought or oversold, indicating when the price is unreasonably high or low, which could suggest a trend reversal. Examples of oscillators include the Stochastic Oscillator, Relative Strength Index (RSI), and Money Flow Index.
\item \textbf{Band systems:} These assist in determining whether prices are overbought or oversold on a relative basis. They use upper and lower bands in conjunction with a moving average. When there is no defined trend, these indicators give a sell signal when the price is above the upper band and a buy signal when the price falls below the lower band. Examples of band systems include the Keltner Channel and Bollinger Bands.
\item \textbf{Divergence Identifier:} Divergence occurs when the price of an asset moves in the opposite direction of a technical indicator, such as an oscillator. Divergence warns that the current price trend may be weakening and, in some cases, may lead to a reversal in the price direction. Examples of divergence identifiers include the Commodity Channel Index, On Balance Volume, and Moving Average Convergence Divergence.
\end{itemize}

Below, we will better describe a few of these indicators, some of which will be provided to our learning algorithms as extra features in the data.

\subsection{Simple Moving Average}

\begin{figure}
  \centering
  \includegraphics[width=0.5\textwidth]{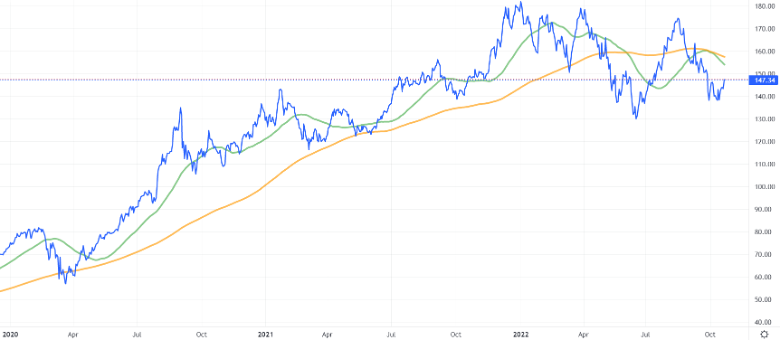}
  \caption{SMA 50 (green) and SMA 200 (orange) in Apple’s stock AAPL instrument.}
  \label{fig:sma}
\end{figure}

The Simple Moving Average (SMA) is calculated by summing up the prices of an asset over a specified number of time periods and dividing this sum by the number of periods. The SMA provides insights into the average price of an asset over a given time frame, smoothing out short-term fluctuations and highlighting underlying trends (Figure~\ref{fig:sma}).

SMA is commonly used to identify current price trends and potential changes in established trends. Short-term SMAs, such as 10-day moving averages, react quickly to price changes, providing timely signals. On the other hand, medium or long-term SMAs, such as 50-day and 200-day moving averages, are slower to react, offering a broader perspective on price movements.

Two popular trading patterns associated with SMAs are the \textit{death cross} and the \textit{golden cross}. A death cross occurs when the 50-day SMA crosses below the 200-day SMA, indicating a \textit{bearish} signal and potential further losses. Conversely, a golden cross happens when a short-term SMA crosses above a long-term SMA, suggesting a \textit{bullish} signal and potential additional gains. Volume analysis is often considered alongside these crosses to validate the strength of the signal.

\subsection{Exponential Moving Average}\label{subsect:ema}

The Exponential Moving Average (EMA) is a type of weighted moving average that assigns more weight to recent price data. The EMA, like the SMA, helps identify price trends over time, but it aims to improve upon the SMA by placing greater emphasis on recent prices, which are considered more relevant than older data. Due to its weighting scheme, the EMA reacts faster to price changes compared to the SMA. Similar to the SMA, the EMA can be used in conjunction with trading patterns such as the death cross and the golden cross.

The EMA is calculated using the following formula:
\[
EMA_t = \text{{price}}_t \times W + EMA_{t-1} \times (1-W)
\]
where:
\begin{itemize}
\item \(\text{{price}}_t\) represents the price of an asset at period \(t\).
\item \(EMA_{t-1}\) is the EMA value from the previous period.
\item \(W\) is the weighting factor calculated as \(2/(n+1)\), where \(n\) is the total number of periods.
\item \(EMA_0\) is typically set as the initial value of the simple moving average (SMA) with a window size of \(n\).
\end{itemize}

\subsection{Relative Strength Index}\label{subsect:rsi}

Conceived by J. Welles Wilder Jr. and delineated in his seminal publication~\cite{wilder1978new}, the Relative Strength Index (RSI) is a popular momentum oscillator used to assess overbought or oversold conditions and identify potential trend reversals or corrective price pullbacks.

The RSI quantifies the velocity and magnitude of recent price oscillations in an instrument, juxtaposing the instrument's vigor during bullish days against its resilience during bearish ones. By relating this comparison to price action, traders gain insight into how an instrument may perform.

\begin{figure}
  \centering
  \includegraphics[width=0.5\textwidth]{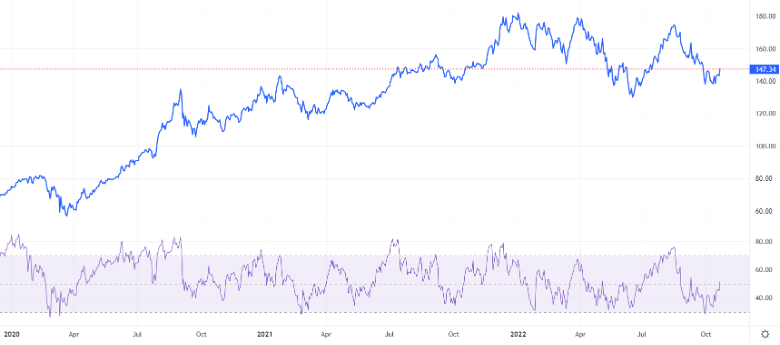}
  \caption{RSI in Apple's stock AAPL instrument.}
  \label{fig:rsi}
\end{figure}

The RSI is represented as a line graph oscillator with a scale ranging from 0 to 100 (Figure~\ref{fig:rsi}). A reading of 70 or higher historically indicates an overbought condition, suggesting that the instrument may be due for a price decline. Conversely, a reading of 30 or lower indicates an oversold market, indicating a potential price increase.

The RSI is calculated using the following formula:

\[
RSI = 100 - \left( \frac{100}{1 + \left( \frac{\text{Average Gain}}{\text{Average Loss}} \right)} \right)
\]

In this calculation, the average gain or loss represents the average percentage gain or loss over a given time period. Periods with price losses are counted as zero in the average gain calculation, and periods with price increases are counted as zero in the average loss calculation. The standard period length for calculating the RSI value is 14.

\section{Related work} \label{sec:relatedWork}

Market forecasting has witnessed extensive exploration of GP, with several studies showcasing its effectiveness. In one of the pioneering efforts, Neely C.~\cite{Neely1997} harnessed GP, aiming to carve out profitability within the forex markets. By analyzing data from six currency pairs between 1981 and 1995, significant improvements were observed in all pairs except the Deutsche mark/yen, when compared to traditional technical analysis rules. Interestingly, the study found that the generated rules did not perform as well on other pairs, even if they involved the same currency.

Iba and Sasaki~\cite{ibasasaki} employed GP to determine optimal Japanese stock companies for investment, as well as the timing and quantities for buying or selling shares. Their comparison of GP with an artificial neural network (ANN) approach revealed that GP outperformed ANN, which tended to suffer from overfitting and yielded inferior results.

Kaboudan~\cite{Kaboudan2000} introduced a metric to quantify the predictability of time series using GP, highlighting the predictability of stock markets. Based on this finding, the author proposed a model that forecasted the high and low prices of the next trading day for six US stocks (Citigroup, Compaq Computers, General Electric, Pepsi, Sears, and Microsoft) and used these predictions to guide trading decisions. The results showed that trading decisions based on GP forecasts achieved higher returns on investment (ROI) compared to those based on ANN forecasts, with GP outperforming ANN in five out of the six stocks.

Potvin et al~\cite{Potvin2004} used GP to generate trading rules in the Canadian stock market, considering data from various industries. Their analysis revealed that GP-generated trading rules proved advantageous during falling or stable market conditions, while the buy and hold approach outperformed GP during rising markets.

Michell and Kristjanpoller~\cite{Michell2020} adopted Strongly-Typed Genetic Programming (STGP) to generate trading rules. They proposed a fitness function that directly evaluated the quality of the generated signals based on their predictive capabilities. The authors used a risk-free metric (rf) to determine buy, sell, and hold signals, considering the stock's return relative to rf. The rules generated by their approach were straightforward and easily interpretable by investors, resulting in higher profits compared to classical optimization approaches and the buy and hold strategy.

Grosan et al~\cite{Grosan2005} and Grosan and Abraham ~\cite{Grosan2006} conducted comparative studies between a genetic multi-expression programming algorithm, ANN, and support vector machine (SVM) models using multiple stock indices for forecasting purposes. The results of both studies strongly supported the use of the genetic model for accurate predictions.
 
Lee and Tong~\cite{Lee2011} combined GP with an ARMA statistical model and ANN to forecast the US GDP, yielding promising results.

Karatahansopoulos et al~\cite{Karatahansopoulos2014} applied GP and gene expression programming (GEP) to achieve profits while trading the Greek ASE20 index. Their findings demonstrated that GP and GEP significantly outperformed other machine learning and trading analysis techniques.

Venturing into automated investment systems, Pimenta et al~\cite{Pimenta2018} used multi-optimization STGP approach. Each individual in the system consisted of two program trees—one for the buy signal and another for the sell signal—which were combined to generate the final trading decision. The authors optimized the algorithm to generate rules that aimed to maximize profits while avoiding overly complex trees. By using the depth of the trees as a complexity metric and removing outliers from the data, their approach achieved returns well above the stock price variation over the same period, surpassing the performance of the other two automated investment strategies tested. Notably, this system generated significant profits even during periods of severe asset depreciation.

Ding et al~\cite{Ding2020} aimed to generate forecasting models using GP to determine the predictability of stock markets. They explored model accuracy through better specification without introducing new variables. The authors argued that different countries' stock markets should have distinct forecasting models and evaluated stock data from China, India, the US, and Japan. Their findings indicated that GP generated approximately three times more profit in developed countries' stocks compared to developing countries.

Ding et al~\cite{Ding2020}, in their quest to decipher market predictability, employed GP to construct forecasting archetypes. They sought to improve model accuracy through better model specification without introducing additional variables. Furthermore, they argued that stock markets in different countries should have distinct forecasting models. Analyzing stock data from China, India, the US, and Japan, with the US and Japan considered as developed countries and China and India as developing countries, the authors generated models for each group. Interestingly, they discovered that GP generated approximately three times more profit in developed countries' stocks than in developing countries.

These diverse studies collectively highlight the potential of GP and its variants in effectively forecasting markets and generating profitable trading strategies across various financial domains and countries.

\section{Data and Methods} \label{sect:data_methods}

\begin{figure}[b]
\includegraphics[width=0.45\textwidth]{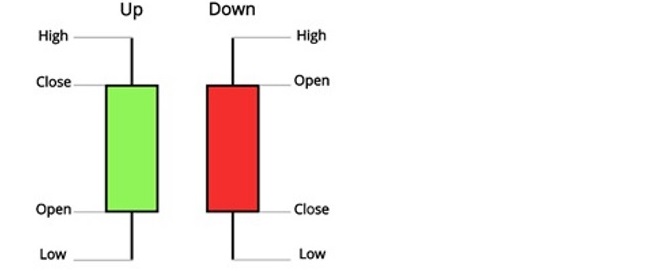}
\\~
\\
\includegraphics[width=0.45\textwidth]{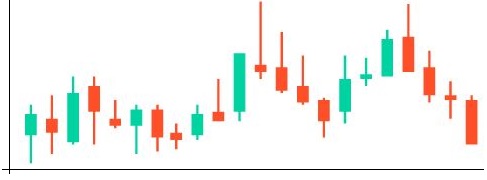}
\caption{Candlestick chart showing the temporal evolution of prices, with detail of two candlesticks showing trends (up, down) and cardinal values (open, close, high, low).}
\label{fig:candle_chart}
\end{figure}

Candlestick charts (Figure~\ref{fig:candle_chart}) are widely used for data visualization in technical analysis. These charts depict price variations over set periods like days, weeks, or months. A single candlestick represents the price change for a specific period, characterized by a rectangular body and extended lines, known as shadows or wicks. The rectangle indicates the opening and closing prices, while the shadows mark the range of prices for that period. A rise in price from the opening to closing is often shown in green or white, indicating a positive (bullish) trend. Conversely, a decline is usually marked in red or black, indicating a negative (bearish) trend.

We used three financial instruments: COTY (Beauty Industry), KO (Food Industry), and PSI20 (Portuguese Index), spanning from January 1, 2015, to April 14, 2022, where each data point refers to one day. These instruments were chosen for their varied market behaviors: COTY shows a declining market trend, KO exhibits a rising market with some price drops, and PSI20 does not strongly show any particular trend. 

\subsection{Input Data Processing}\label{subsect:input_data}

In addition to the four fundamental features (open, close, high, low) and the volume of the asset, we enriched the datasets with some technical indicators and tailored inputs. 

We incorporated Exponential Moving Averages (EMAs, see Section~\ref{subsect:ema}) for several periods (5, 13, 50, and 200 days). This range not only offers varying degrees of data smoothing but also helps agents identify trends spanning short to long durations. The selection of these EMA periods was based on insights from technical trading discussions and resources frequently consulted by traders. We also added the Relative Strength Index (RSI, see Section~\ref{subsect:rsi}), which offers agents an insight into the momentum of instrument price changes. 

Additionaly, we included the 'profit percentage' metric, a feature built dynamically in runtime that acts as a real-time gauge, allowing agents to evaluate the success of their current trades and adapt accordingly. Initially (on the first row of the data being used for calculating fitness), and every time the agent does not have a buy or sell position (more on this later), the profit percentage is zero. From that point on, the calculation is made as follows: $\text{profit percentage} = \left( \text{Profit}/{M} \right) \times 100$, where Profit is calculated differently depending on whether the agent currently has a buy or sell position: if buy, $\text{Profit} = \text{SharesNumber} \times C - M$, and if sell, $\text{Profit} = M - \text{SharesNumber} \times C$, where M is the amount of money used to trade (in our case, always set to 1000), C is the close price, and SharesNumber is the number of shares that can be bought with money M, considering the open price.

As tailored features, we introduced 'smallEmaDiff' and 'bigEmaDiff,' both derived from EMAs. 'SmallEmaDiff' represents the difference between the 5-day and 13-day EMAs, whereas 'bigEmaDiff' captures the gap between the 50-day and 200-day EMAs. These metrics assist agents in spotting potential trend shifts.

By adding four EMAs, the RSI and the profit percentage, and our two tailored metrics to the five initial features, we provide agents with 13 features that should provide a rounded view of the market.

\subsection{Output Data and Fitness}

For each row of the input data, the evolving agent (the individual in the population) produces an output that is interpreted as a buy or sell signal (or a ``hold'' decision that is a state of inactivity). The output produced may take different forms, depending on which GP variant is being used, and in some cases requires some processing before being interpreted (explained later).

Regarding the fitness function, we initially selected the return on investment (ROI) as the metric to optimize. The ROI is calculated as a percentage, $\text{ROI} = 100 \times \text{TotalProfit} / \text{InitialInvestment}$, where TotalProfit is the net profit calculated on all the (sequential) rows of the dataset and InitialInvestment is the initial amount of money (1000) invested. Although the ROI offers a direct indication of the returns an agent derives from its decisions, using it as fitness revealed certain challenges. A notable concern was the inclination of some agents to abstain from trading. While these agents might seem to showcase high fitness due to the absence of losses, this inactivity contradicts the primary objective of a trading strategy, which is to actively participate in trading. To counter this, we imposed a penalty: agents that remained inactive were attributed the worst possible fitness, thus minimizing their chances of being selected. However, penalizing inactivity was not the complete solution. It was vital to incentivize agents that actively traded and exhibited discernment in their decisions. Therefore, we modified our fitness assessment by incorporating also the win rate (proportion of transactions with profit), such that fitness is the product of ROI and win rate. This combined score accentuates the rewards for agents boasting a high win rate. However, this is only considered when the ROI is positive. Multiplying the win rate by negative ROI scores would unintentionally diminish loss magnitudes, deviating from our intent of nurturing agents that trade actively and wisely.

Therefore, the fitness function is the ROI, multiplied by the win rate when the ROI is positive. Any fitness above zero means a profitable strategy. As seen later, the magnitude of the fitness is very different between the training and test data. This happens not only because it is difficult that a learned strategy generalizes well in unseen data, but also because the test set is smaller, providing a shorter time span to reach high profit.

\subsection{Standard and Vectorial GP} \label{subsect:sdt_vgp}

The most distinctive characteristic between standard GP and VGP is that standard GP relies on scalar features, whereas in VGP the features are vectors. Unlike the original VGP~\cite{azzali2019vectorial}, which could manage vectors of varying sizes, all vectors in our problem are restricted to a size of either 1, for scalar values, or 21, representing sequences of 21 working days ($\approx$4 weeks, since traditional market closes on weekends). We deviate from the original VGP formulation also in that we do not use exactly the same function set. For example, we leave out functions like max, min, mode, length, norm, and others, as well as the parametric aggregate functions, while introducing others that were not used in the original VGP, like the standard deviation of a vector. For standard GP, we use basic arithmetic operations, some trigonometric functions and a couple of constant-generating tailored functions. Tables~\ref{table:SGP} and~\ref{table:VGP} provide details on the function sets used in standard GP and VGP, respectively. Note that, in vectorial operations of arity 2, if one of the arguments is a scalar and the other is a 21-dimensional vector, the scalar is replicated 21 times to form another 21-dimensional vector.

In standard GP, the output produced by the agent on each row of the dataset is a scalar, interpreted like this:
\begin{itemize}
\item A value equal or above 1 is a buying signal.
\item A value equal or below -1 is a selling signal.
\item Any other value means a state of inactivity\\ (a ``hold'' decision from the agent).
\end{itemize}
In VGP, if the output is a vector of size 1, the single value is interpreted the same way; if it is a vector of size 21, the average of the 21 values is calculated and then also interpreted the same way.

\begin{table*}
\centering
\caption{Function set of standard GP.}
\begin{tabular}{rll}
\textbf{Name} & \textbf{Description} & \textbf{Example} \\
\grayline
ADD & Adds two input variables & $1 + 2 = 3$ \\
\grayline
MULT & Multiplies two input variables & $2 \times 2 = 4$ \\
\grayline
SUB & Subtracts two input variables & $2 - 1 = 1$ \\
\grayline
DIV & Divides two input variables, returns 1 if denominator is 0, as in~\cite{koza1992} & $10 / 5 = 2$ \\
\grayline
NEG & Negates a single input variable & NEG$(10) = -10$ \\
\grayline
SIN & Returns the sine of a single input variable & SIN$(0) = 0$ \\
\grayline
COS & Returns the cosine of a single input variable & COS$(0) = 1$ \\
\grayline
TAN & Returns the tangent of the argument & TAN$(0) = 0$ \\
\grayline
SIGNUM & Returns 1(-1) if variable is positive(negative), 0 otherwise & SIGNUM$(10) = 1$ \\
\grayline
GT & Returns 1 if its first input variable is greater than its second input variable, -1 otherwise & GT(10, 5) $= 1$ \\
\hline
\end{tabular}
\label{table:SGP}
\end{table*}

\begin{table*}
\centering
\caption{Function set of vectorial GP.}
\begin{tabular}{rp{8cm}l}
\textbf{Name} & \textbf{Description} & \textbf{Example} \\
\hline
ADD & Adds the corresponding elements of two vectors & [1, 2, 3] $+$ [4, 5, 6] = [5, 7, 9] \\
\grayline
MULT & Multiplies the corresponding elements of two vectors & [1, 2, 3] $\times$ [4, 5, 6] = [4, 10, 18] \\
\grayline
SUB & Subtracts the corresponding elements of two vectors & [4, 5, 6] $-$ [1, 2, 3] = [3, 3, 3] \\
\grayline
DIV & Divides~\cite{koza1992} the corresponding elements of two vectors & [4, 9, 8] $/$ [2, 3, 0] = [2, 3, 1] \\
\grayline
DOT & Calculates the dot product of two vectors & [1, 2, 3] $\cdot$ [4, 5, 6] = 32 \\
\grayline
NEG & Negates the elements of a vector & NEG$([-1, 2, -3]) = [1, -2, 3]$ \\
\grayline
SIN & Calculates the sine of each element in a vector & SIN$([0, \pi/2, \pi]) = [0, 1, 0]$ \\
\grayline
COS & Calculates the cosine of each element in a vector & COS$([0, \pi/2, \pi]) = [1, 0, -1]$ \\
\grayline
TAN & Calculates the tangent of each element in a vector & TAN$([0, \pi/4]) = [0, 1]$ \\
\grayline
MEAN & Calculates the mean of a vector & MEAN$([1, 2, 3, 4]) = 5$ \\
\grayline
STD\_VAR & Calculates the standard deviation of a vector & STD\_VAR$([1, 2, 3, 4]) = 1.29$ \\
\grayline
CUM\_MEAN & Calculates the cumulative mean of a vector & CUM\_MEAN$([1, 2, 3, 4]) = [1, 1.5, 2, 2.5]$ \\
\grayline
GT\_THAN & Returns 1(-1) if the mean of the first variable is greater(lower) than the mean of the second variable, 0 if the means are equal & GT\_THAN$([5, 3],[1, 3]) = 1$\\
\hline

\end{tabular}
\label{table:VGP}
\end{table*}

\subsection{Complex Vectorial GP}

Complex VGP (CVGP) extends the VGP method by enabling the use of complex numbers in the evolved expressions. Complex numbers contain both a real and an imaginary component, which allows the agents to perform operations that are infeasible when relying solely on real numbers. For instance, the logarithm or the square root of a negative real number, previously requiring special protected functions, now become valid operations. Although this increases the size of the search space, it also provides more space for exploring solutions.

Similar to the VGP method, CVGP accepts vectorial inputs. Likewise, each vector can be unidimensional or possess a dimensionality of 21, but now the vectors are made of complex numbers. Table~\ref{table:CVGP} details the function set of CVGP. Also the outputs of the agents are complex numbers. To interpret them as buy/sell signals, we use only the real part of the numbers, everything else like in VGP.

\begin{table*}
\centering
\caption{Function set of complex vectorial GP.}
\begin{tabular}{rp{6.5cm}l}
\textbf{Name} & \textbf{Description} & \textbf{Example} \\
\hline
ADD & Element-wise addition of two complex vectors & [1+2i, 3-4i] $+$ [5-i, 2+3i] = [6+i, 5-i] \\
\grayline
MULT & Element-wise multiplication of two complex vectors & [1+2i, 3-4i] $\times$ [5-i, 2+3i] = [7+9i, 18+i] \\
\grayline
SUB & Element-wise subtraction of two complex vectors & [1+2i, 3-4i] $-$ [5-i, 2+3i] = [-4+3i, 1-7i] \\
\grayline
DIV & Element-wise division of two complex vectors & [1+2i, 3-4i] $/$ [5-i, 2+3i]) = [0.115+0.423i, -0.462-1.308i] \\
\grayline
DOT & Dot product of two complex vectors & [1+2i, 3-4i] $\cdot$ [5-i, 2+3i] = -3+6i \\
\grayline
NEG & Negation of a complex vector & NEG([1+2i, 3-4i]) = [-1-2i,-3+4i] \\
\grayline
LOG & Natural logarithm of a complex vector & LOG([1+2i, 3-4i]) = [0.805+1.107i,1.609-0.927i] \\
\grayline
SQRT & Square root of a complex vector & SQRT([1+2i, 3-4i]) = [1.272+0.786i, 2-i] \\
\grayline
SIN & Sine of a complex vector & SIN([1+2i, 3-4i]) = [3.166+1.960i,3.854+27.017i] \\
\grayline
COS & Cosine of a complex vector & COS([1+2i, 3-4i]) = [-2.033-3.052i,-27.035+3.851i] \\
\grayline
TAN & Tangent of a complex vector & TAN([1+2i, 3-4i]) = [0.034+1.015i,-0.0002-0.999i] \\
\grayline
MEAN & Mean of a complex vector & MEAN([1+2i, 3-4i, 5-7i]) = 3-3i \\
\grayline
CUM\_MEAN & Cumulative mean of a complex vector & CUM\_MEAN([1+2i, 3-4i, 5-7i]) = [1+2i,2-i,3-3i] \\
\grayline
GT\_THAN\_ REAL & Returns 1 if the real part mean of the first variable is greater than the real part mean of the second variable, otherwise returns -1 & GT\_THAN\_REAL([1+2i, 3-4i], [5-i, 2+3i]) = [-1, 0i] \\
\grayline
GT\_THAN\_ COMPLEX & Returns i if the imaginary part mean of the first variable is greater than the imaginary part mean of the second variable, otherwise returns -i  & GT\_THAN\_COMPLEX([1+2i, 3-4i], [5-i, 2+3i]) = [0,-i] \\
\hline
\end{tabular}
\label{table:CVGP}
\end{table*}

\subsection{Strongly-Typed Vectorial GP}

Strongly-Typed Genetic Programming (STGP) introduces a type system to standard GP, dictating input and output types for functions and terminals and ensuring syntactic correctness in the resulting solutions. Likewise, Strongly-Typed Vectorial GP (STVGP) extends VGP by integrating boolean and vector value types. The aim is to provide rule-like strategies that improve interpretability and invite the experts to understand the reasoning behind the decisions and eventually to incorporate their own knowledge, potentially improving model performance.
STVGP uses three main function categories:
\begin{itemize}
\item Vectorial functions that accept and produce vectors, enabling data structure manipulation
\item Relational functions that use vectors and/or booleans, outputting boolean values, and aid in modeling relationships and conditions
\item Boolean functions that operate with boolean inputs and outputs, providing logical operations and decision-making capabilities.
\end{itemize}
In STVGP, the initial node of a strategy is a relational or a boolean function, and the agent outputs a boolean value that is interpreted as buy (true) or sell (false). Table~\ref{table:STVGP} contains the details of the function set used by STVGP, except the functions that were already part of VGP (Table~\ref{table:VGP}). Note that function GT\_MEAN of VGP, previously outputting 1 or -1, now outputs true or false.

\begin{table*}
\centering
\caption{Function set of strongly-typed vectorial GP, except the functions already in Table~\ref{table:VGP}.}\vspace{-5pt}
\begin{tabular}{rp{9cm}l}
\textbf{Name} & \textbf{Description} & \textbf{Example} \\
\hline
SUM\_GT & Returns true(false) if the sum of the first variable is greater(lower) than the sum of the second variable & SUM\_GT$([1, 2, 3, 4], [1, 1, 2, 3])$ = false \\
\grayline
AND & Returns true if all input boolean values are true, false otherwise & AND$($true, false$) =$ false \\
\grayline
OR & Returns true if at least one input boolean value is true, false otherwise & OR$($false, true$) =$ true \\
\grayline
XOR & Returns true if exactly one input boolean value is true, false otherwise & XOR$($true, true$) =$ false \\
\grayline
NOT & Returns the inverse of the input boolean value & NOT$($true$) =$ false \\
\grayline
IF\_ELSE & Returns the result of a ternary operator & IF\_ELSE$($true, [1, 2, 3], [4, 5, 6]$) = $[1, 2, 3] \\
\hline
\end{tabular}
\label{table:STVGP}
\end{table*}

\section{Experimental Setup and Overfitting}

\begin{figure*}
\includegraphics[width=0.31\textwidth]{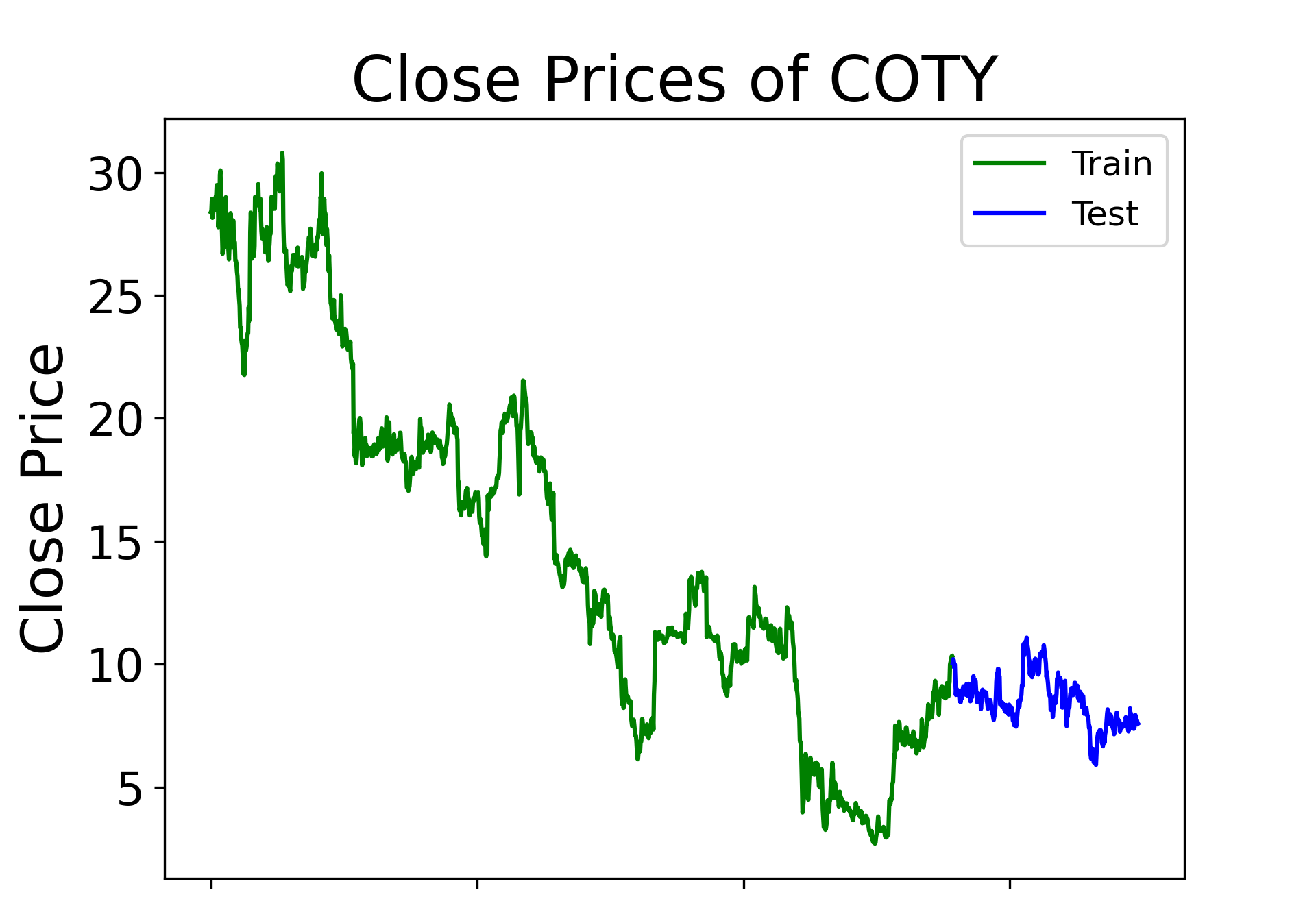}
\hfill
\includegraphics[width=0.31\textwidth]{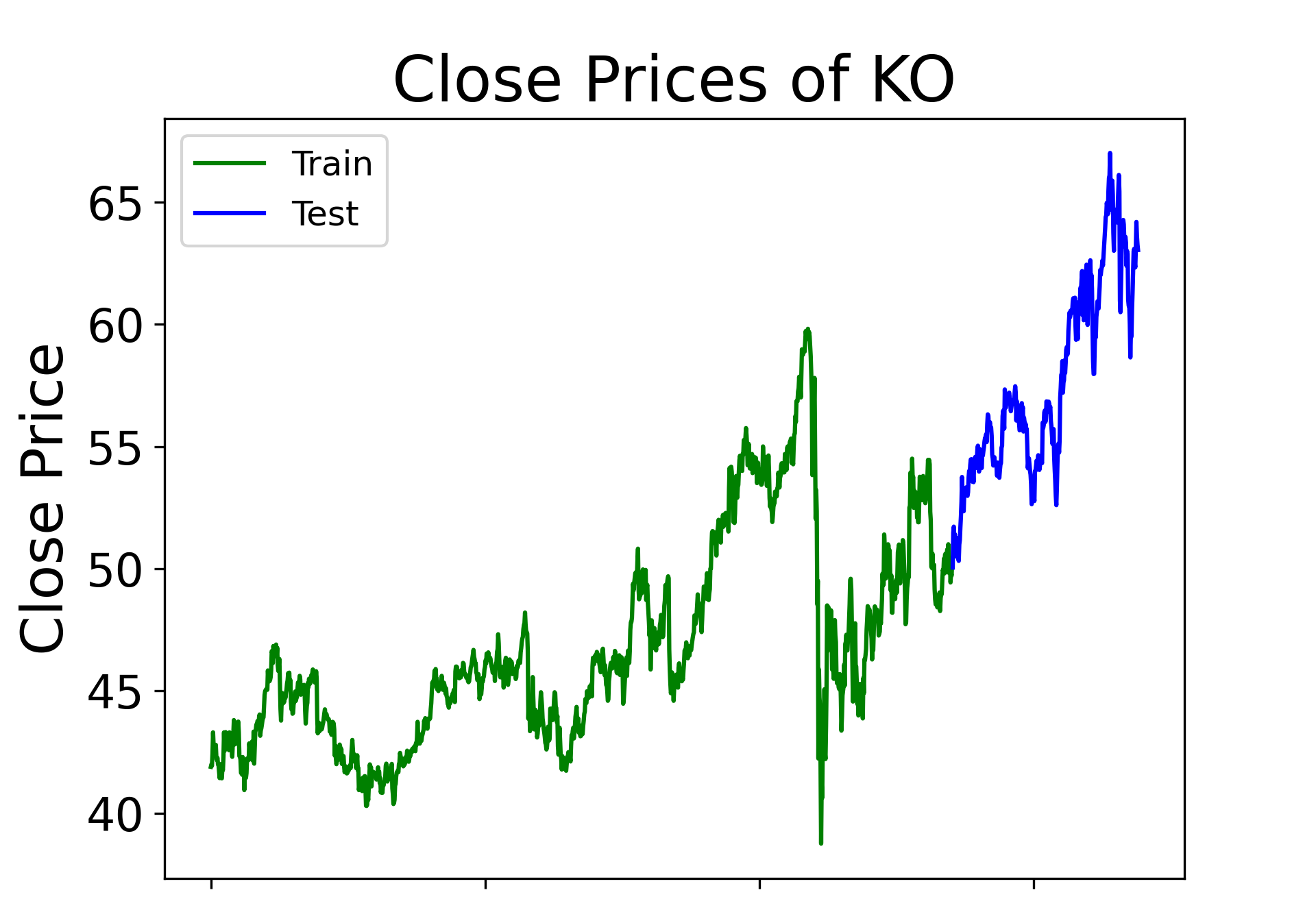}
\hfill
\includegraphics[width=0.30\textwidth]{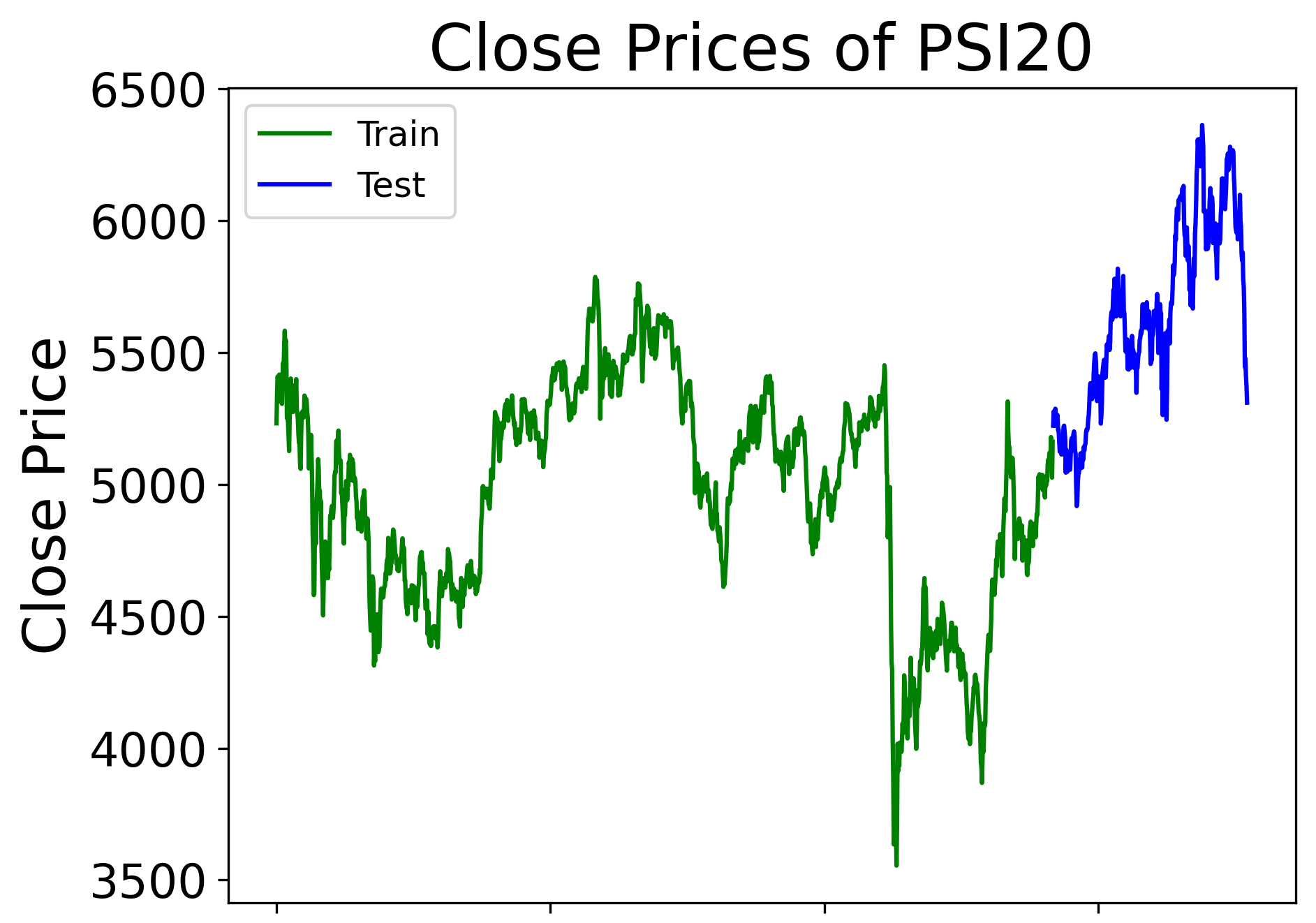}
\vspace{-5pt}\caption{Instrument close prices spanning January 1, 2015, to April 14, 2022.}
\label{fig:data}
\end{figure*}

Once enriched with the added features (Section~\ref{subsect:input_data}), the three datasets (COTY, KO, PSI20) were divided using an 80-20 split for training and testing. Given the time-series nature of our data, we partitioned it sequentially without shuffling. Visual representations of one of the features (close prices) are provided in Figure~\ref{fig:data}, where training and test splits are shown in green and blue, respectively.

All the GP variants were implemented in the Jenetics library~\cite{wilhelmstotterjenetics} and preliminary tests were performed to determine a few of the running parameters, all others remaining with the default setting. We used populations of 3000 individuals, allowed to evolve for 50 generations. Selection was performed with tournaments of size 30 (1\% of the population) and the mutation rate was 0.001. Due to computational time constraints, we did only 10 runs per experiment, instead of the usual 30. Depth and size limits were imposed (13 and 90, respectively) on the evolving trees, to avoid bloat.

To mitigate overfitting, we added a substantial degree of randomness to the training data. Instead of always considering the entire series from the first to the last day, we use a buffer of 100 days ($\approx$5 months) at the beginning of the time series where, for each generation, the first day is chosen randomly, while the last day is chosen such that the different training sets all have the same size. Furthermore, after establishing the training set for a generation, this data is segmented into three equal parts, an approach inspired by the work of Gonçalves and Silva~\cite{Goncalves2013}. In most generations, agents are evaluated in only one part, while during 'super generations', specifically those divisible by 50, agents are provided access to all three parts.

\section{Results} \label{sect:results}

\begin{figure*}
\centering
\begin{tabular}{@{}c@{}c@{}c@{}}
\textbf{COTY}&\textbf{KO}&\textbf{PSI20}\\
\includegraphics[width=0.33\textwidth]{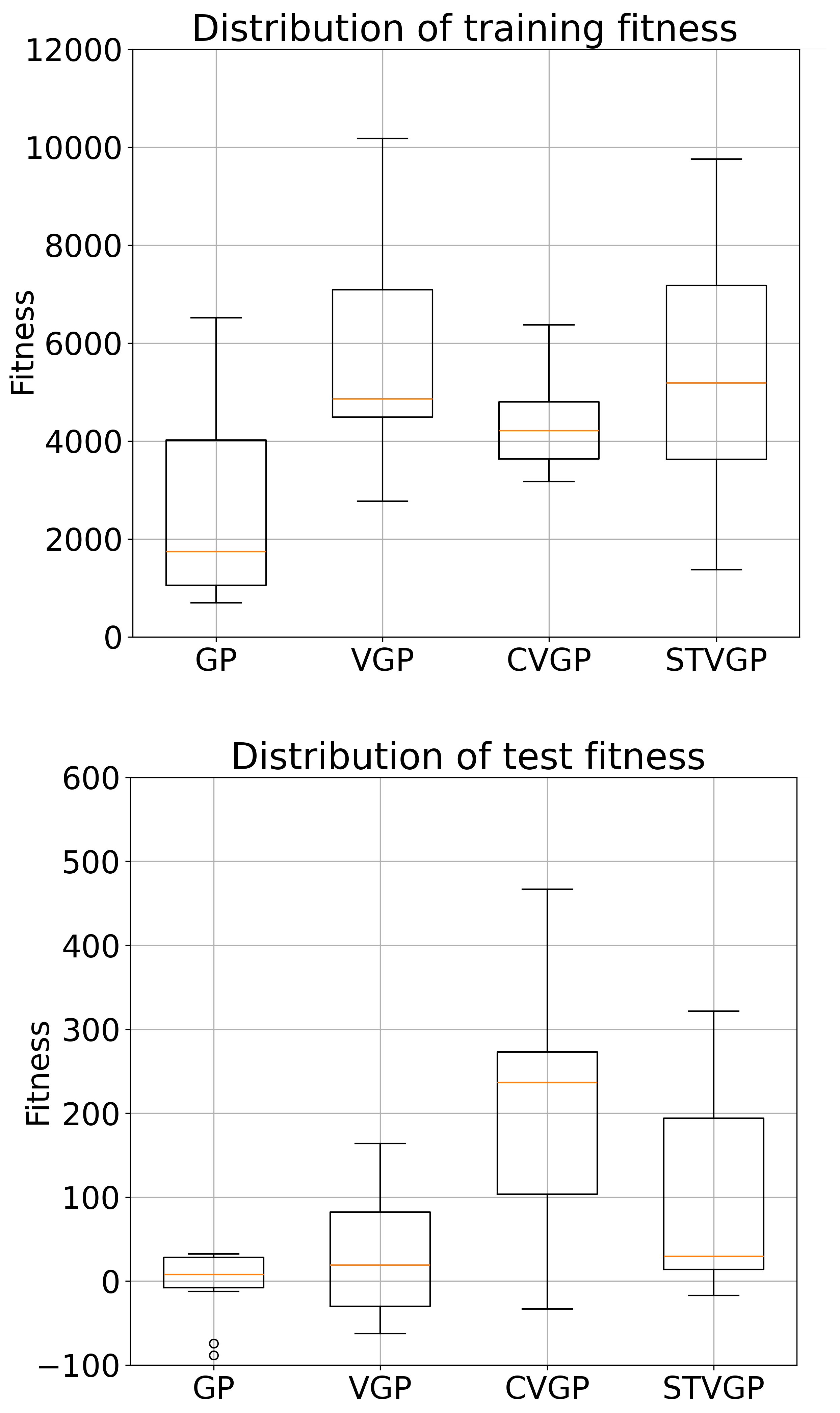}
&
\includegraphics[width=0.33\textwidth]{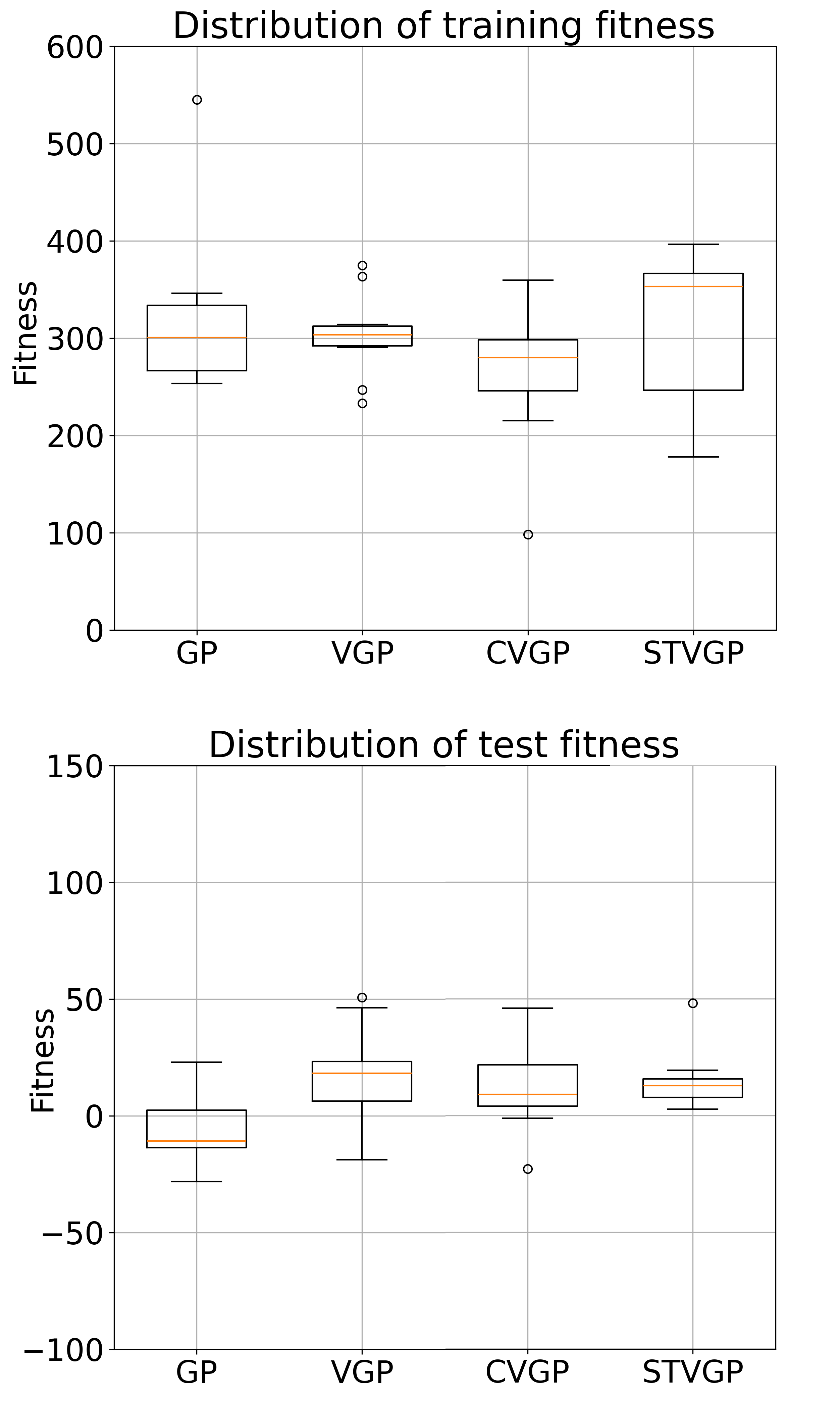}
&
\includegraphics[width=0.33\textwidth]{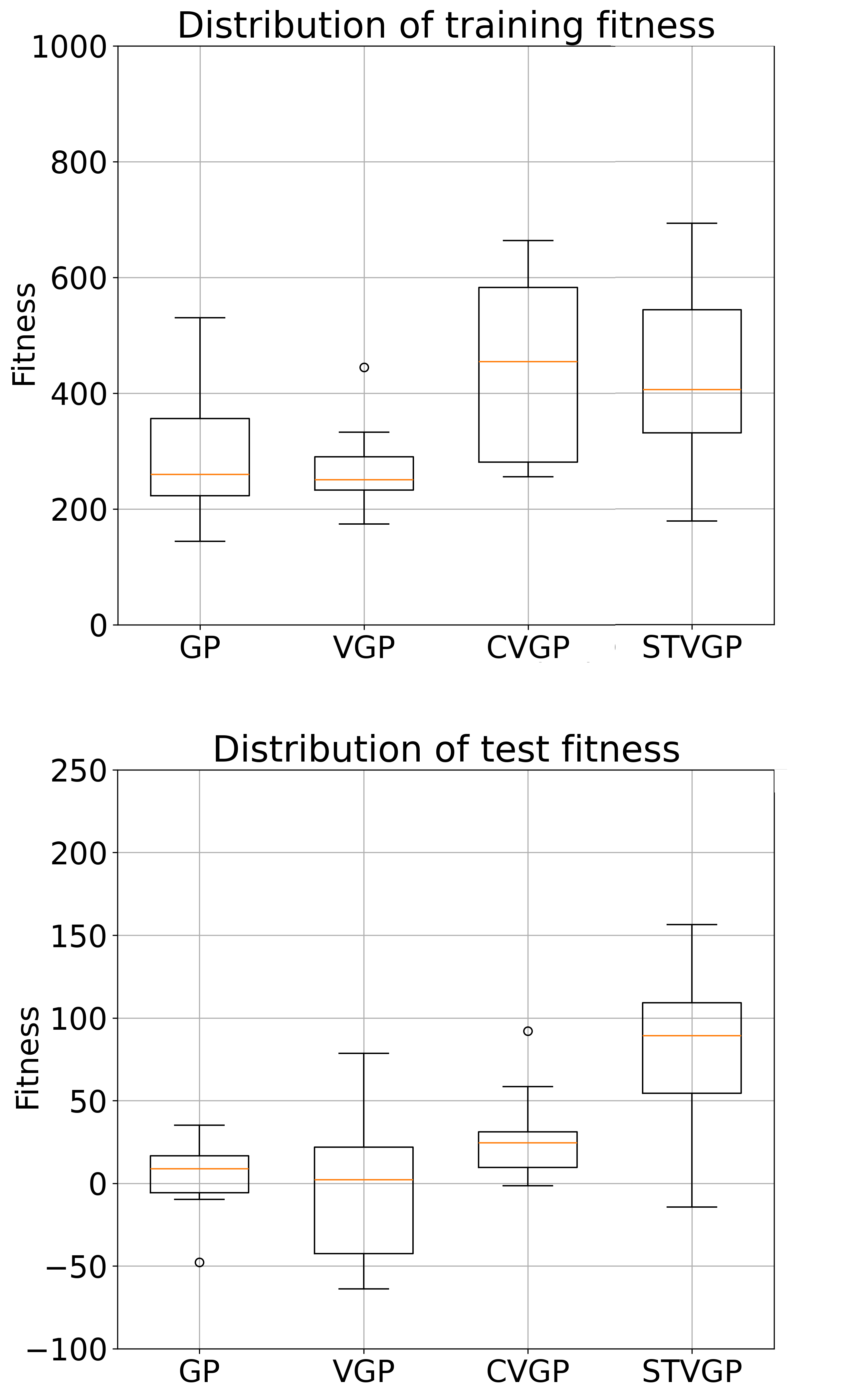}
\end{tabular}
\caption{Fitness distribution in the last generation on training (top row) and test (bottom row) for each dataset.}
\label{fig:boxplots}
\end{figure*}

The interpretation of the obtained results is very straightforward. Figure~\ref{fig:boxplots} shows the boxplots of training and test fitness in the last generation, while Table~\ref{tab:pvalues} shows the $p$-values of the pairwise comparisons of test fitness (according to the Kruskal-Wallis non-parametric test with significance level 0.05). 

\begin{table*}
\caption{Pairwise $p$-values obtained with Kruskal-Wallis. For each dataset, \color{green}green\color{black}/\color{red}red \color{black} means the method listed in the first column is significantly better/worse than the method listed in the first row, with significance level 0.05.}
\small
\centering
\begin{tabular}{ccc}
\begin{tabular}{|c|c@{\hspace{6pt}}c@{\hspace{6pt}}c@{\hspace{6pt}}c|}
\hline
\textbf{COTY}&GP&VGP&CVGP&STVGP\\
\hline
GP &---&0.2697&\color{red}{0.0092}&0.0576\\
VGP &0.2697&---&\color{red}{0.0193}&0.2004\\
CVGP    &\color{green}{0.0092}&\color{green}{0.0193}&---&0.2332\\
STVGP &0.0576&0.2004&0.2332&---\\
\hline
\end{tabular}
&
\begin{tabular}{|c|c@{\hspace{6pt}}c@{\hspace{6pt}}c@{\hspace{6pt}}c|}
\hline
\textbf{KO}&GP&VGP&CVGP&STVGP\\
\hline
GP &---&\color{red}{ 0.0009}&\color{red}{0.0023}&\color{red}{0.0013}\\
VGP &\color{green}{ 0.0009}&---&0.4015&0.2332\\
CVGP    &\color{green}{ 0.0023}&0.4015&---&0.5660\\
STVGP &\color{green}{ 0.0013}&0.2332&0.5660&---\\
\hline
\end{tabular}
&
\begin{tabular}{|c|c@{\hspace{6pt}}c@{\hspace{6pt}}c@{\hspace{6pt}}c|}
\hline
\textbf{PSI20}&GP&VGP&CVGP&STVGP\\
\hline
GP &---&0.7573&\color{red}{0.0305}&\color{red}{0.0041}\\
VGP &0.7573&---&0.0851&\color{red}{0.0031}\\
CVGP    &\color{green}{ 0.0305}&0.0851&---&\color{red}{0.0152}\\
STVGP &\color{green}{ 0.0041}&\color{green}{0.0031}&\color{green}{0.0152}&---\\
\hline
\end{tabular}
\end{tabular}
\label{tab:pvalues}
\end{table*}

\begin{figure*}
\centering
\begin{tabular}{ccc}
\textbf{COTY}&\textbf{KO}&\textbf{PSI20}\\
\includegraphics[width=0.22\textwidth]{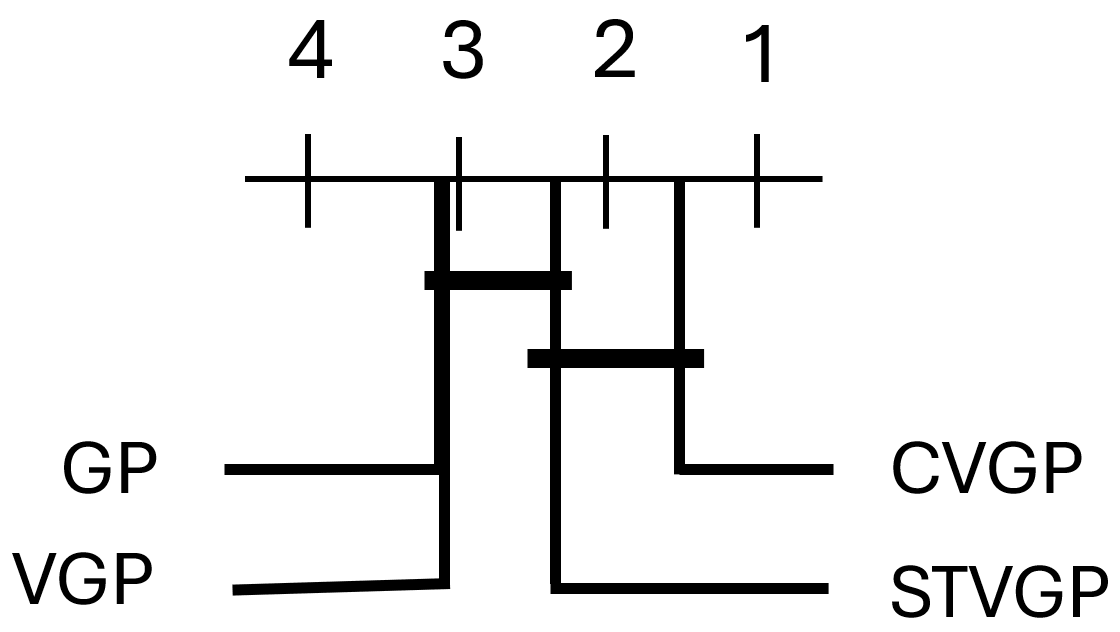}
&
\hspace{30pt}
\includegraphics[width=0.22\textwidth]{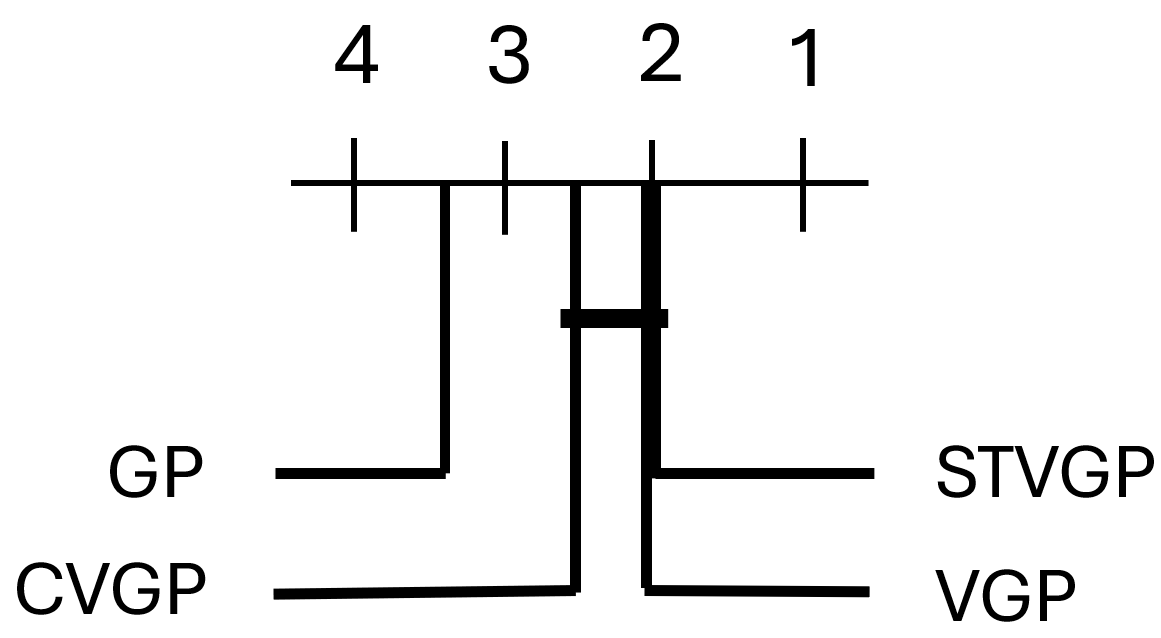}
\hspace{30pt}
&
\includegraphics[width=0.22\textwidth]{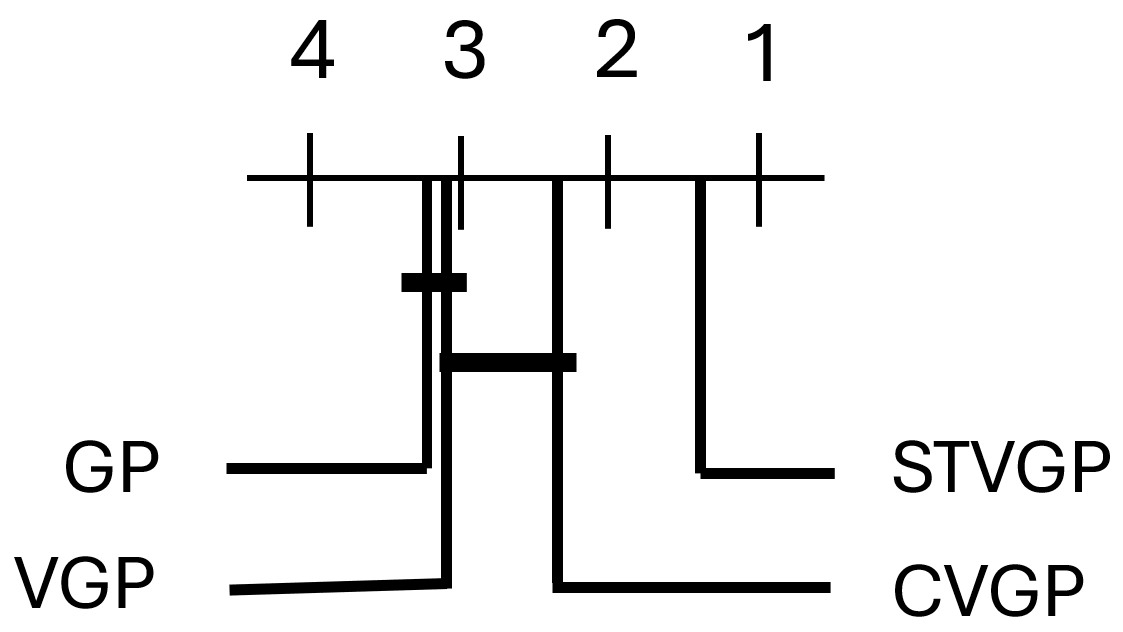}
\end{tabular}
\caption{Critical Difference diagrams for each dataset.}
\label{fig:diagrams}
\end{figure*}

All the methods achieve positive fitness in training, but that is not always the case in test, revealing that generalization is difficult in these datasets. Although the median test fitness is always above zero except in one case (standard GP on KO), the distribution of results frequently includes positive and negative fitness values, with only a few notable cases where both box and whiskers are above zero. Standard GP is often significantly worse than others, whereas CVGP and STVGP are most commonly significantly better than others. We corroborate this in Figure~\ref{fig:diagrams} that shows a Critical Difference (CD) diagram for each dataset. In a CD diagram, the methods are ordered according to their mean rank in all 10 runs. For example, a method that always ranked first would be positioned in 1, while a method that always ranked last would be positioned in 4. A horizontal bar connecting two or more methods means there is no statistically significant difference between them. As we can see, standard GP is always one of the worst methods while STVGP is always among the best.

Although we do not disclose the strategies of the best individuals, we report that they abundantly use medium-term price trends (e.g., EMA50) and short-term price shifts (smallEmaDiff), trigonometric functions (particularly the tangent) that we hypothesize to fit oscillatory data patterns, the close price (but not the open price), besides basic operations like addition and subtraction. Also frequently used are the functions that produce constants from conditional logic (e.g., SIGNUM, GT\_THAN, GT\_THAN\_REAL), which makes sense for decision-making, and STVGP indeed uses the IF\_ELSE function. The RSI is frequently used by CVGP and STVGP, but not by the other methods. Other, more complex functions and terminals, like dot product, cumulative mean, standard deviation, and profit percentage are less used than expected.

\section{Conclusions and Future Work} \label{sect:conclusions}

We have addressed the difficult task of automatically creating profitable trading strategies. Building on a recent GP formulation that accepts features in the form of vectors, vectorial GP, we have adapted it to the particularities of our data and then developed two other variants: one that deals with complex numbers, complex vectorial GP, and another that introduces types, resulting in strongly-typed vectorial GP. A comparison between standard GP and the three vectorial variants revealed that, on three financial instruments representing different market situations, standard GP was always among the worst while strongly-typed vectorial GP was always among the best. In a few cases, it was possible to consistently evolve profitable strategies. The results obtained by complex vectorial GP were also extremely interesting, which makes us believe that a complex strongly-typed variant should be developed and tested in the future.

Another possible improvement to this line of work would be to evolve different types of agents that work together, each specialized in a given task. For example, one agent could be specialized in managing loss (i.e., would make decisions only when faced with a loss) while another agent could manage profit, since the right decisions for each case will probably depend on different factors.

Naturally, we can also add many other technical indicators for the evolutionary methods to work with, either as extra features in the dataset or as elements of the primitive set. A larger search space demands a larger population and more computational power, but the promising results we achieved in this work suggest that such extra efforts may bring substantial rewards. 

\begin{acks}
This work was partially funded by FCT through the LASIGE Research Unit, ref. UID/00408/2025.
\end{acks}

\balance
\bibliographystyle{ACM-Reference-Format}
\bibliography{menoita}

\end{document}